# A Channelized Binning Method for Extraction of Dominant Color Pixel Value

Siddu P Algur
Department of Information Science
B. V. Bhoomraddi College of
Engineering and Technology
Hubli, Karnataka, INDIA
algursp@bvb.edu

N H Ayachit
Department of Physics
B. V. Bhoomraddi College of
Engineering and Technology
Hubli, Karnataka, INDIA
nhayachit@gmail.com

Vivek R
Department of Information Science
B. V. Bhoomraddi College of
Engineering and Technology
Hubli, Karnataka, INDIA
vivekr321@gmail.com

*Abstract—Color is one of the most important and easily identifiable features for describing the visual content. The MPEG standard has developed a number of descriptor that covers different aspects of the visual content. Dominant color descriptor is one of them. This paper proposes a channelized binning approach, a novel method for extraction of dominant color pixel value, which is a variant of dominant color descriptor. The Channelized binning method treats the problem as statistical problem and tries to avoid color quantization and interpolation – guessing of number and centroid of dominant colors. Channelized binning is an iterative approach which automatically estimates the number of dominant pixel values and their centroids. It operates on 24 bit full RGB color space, by considering one color channel at a time and hence avoiding the color quantization. Results show that the proposed method can successfully extract dominant color pixel values.*

I. INTRODUCTION

Image processing and its application have always kept the research community busy by providing new challenges and revealing new solution in phases. One of the major challenges is to extract data from images.

The way colors are characterized in an image is by relating to its perception, coherency and spatial distribution. Several approaches to its characterization have been proposed. The *MPEG-7* which is a a content representation standard for information search has standardized a subset of these approaches in the form of color descriptors.

The *MPEG-7* defines five color descriptors, as against seven color descriptors but *color space* and *color quantization* are considered as basic block as proposed in [1],[2]. These descriptors cover different aspects of color and application areas. These five color descriptors are Dominant Color, Scalable Color, Color Structure, Color Layout, Group of Frames/Group of Pictures Color.

Dominant color descriptor (DCD) provides a compact description of the representative colors of an image or image region. Since DCD stores only the dominant colors, a maximum of eight colors, instead of a color histogram, the storage requirement is very effective with relatively small redundancy. Its main target applications are similarity retrieval in image databases and browsing of image databases based on single or several color values. In its basic form, the *dominant color* descriptor consists of the number of *dominant colors* (N), and for each *dominant color* its value expressed as a vector of *color components* ($c_i$) and the *percentage of pixels* ($p_i$) in the image or image region in the cluster corresponding to $c_i$. Two additional fields, *spatial coherency* (s) and *color variance* ($v_i$), provide further characteristics of the color distribution in the spatial and color space domains [2].

In order to extract the representative colors, some efforts have focused on the color segmentation by clustering methods. However, true type color images consist of more than 16 million (224) different colors in a 24 bit full RGB color space, therefore the conventional clustering algorithms are very time consuming. Moreover, it is difficult to obtain global optimal for DCD. The GLA algorithm [3] is the most extensively used algorithm to extract the dominant color from an image; however, it needs extensive computation cost. Furthermore, there are three intrinsic problems associated with the existing algorithms such as, (1) It may give quite different kinds of clusters when cluster number is changed. (2) A correct initialization of the cluster centroid is a crucial issue, because some clusters maybe empty if their initial centres lie far from the distribution of data. (3) The effectiveness of the GLA depends on the definition of "distance"; therefore, different initial parameters of an image may cause different clustering results.

The present paper which is based on identifying dominant pixel values is organized as follows: The channelized binning for extracting the dominant color pixel value is explained in section 2. A discussion and comparison of experiments and results are made in section 3. Finally, a short conclusion is presented in section 4.

II. PROPOSED WORK

To overcome the limitations of existing algorithms as described in above section channelized binning method has proposed, this method operates on 24 bit full RGB color. To avoid working on 16 million different colors at once and to avoid color quantization where in number of colors are reduced , each color channel is dealt separately, i.e. red channel represented by 8 bits, green channel represented by 8 bits and similarly blue channel. By doing so, the obtained result will consists of dominant pixel values and their percentage of occurrence for all three channels separately.

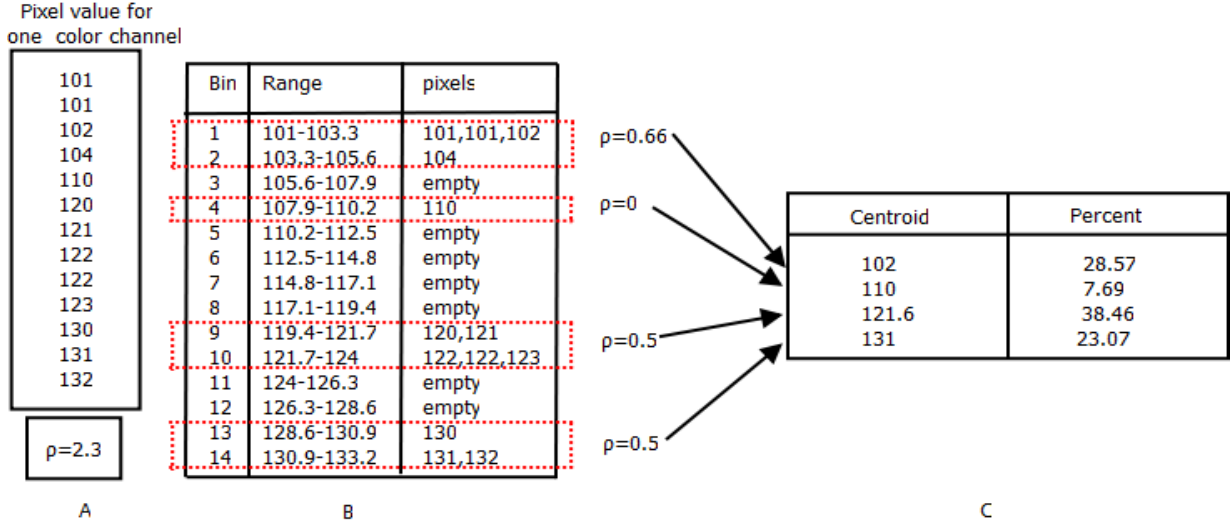
Figure 1: Example 1 calculation on hypothetical pixel values

The proposed method initially separates the pixel value into three channels namely R, G, and B. The pixel values are then sorted in ascending order of their occurrence as described in Algorithm 1. Later a two phased method is employed as described below.

### A. Phase 1: Channelized Binning

In this phase, the range and centroid of the color bins are estimated and pixel values are binned into different color bins (classified into different color bins). The Algorithm 2 describes the various steps in this phase.

Average sum of difference of pixel values is used as the criteria to classify the color denoted by $'\rho'$. Equation 1 shows the calculation $'\rho'$, where n is the number of pixel present in the image considered.

$$\rho = \frac{\sum_{i}^{n} pixel_i - pixel_{i-1}}{n-1} \quad (1)$$

The value $\rho$ serves the range for each color bin; the first color bin ranges from the least pixel value pixel0 to pixel0+ $\rho$, second color bin ranges from pixel0+ $\rho$ to pixel0+ $2\rho$ and so on till the highest pixel values falls under a color bin. It is important to note that the value of $\rho$ may take value less than or greater than 1, if the value of $\rho$<1, then it may lead to many empty bins. In such cases empty bins are eliminated in further iterations. Experimental results show that always clusters of empty bins occurs together. Consider that ith bin is empty, out of possible n bins. Then the whole process is repeated for pixels falling under 1 to i-1 bins and for pixels falling under i+1 to n bins separately, this is continued until there is no occurrence of empty bins. In this phase the bin size is overestimated leading to smaller $\rho$ value, gradually in each iteration empty bins are eliminated and re-estimating the $\rho$, leading to optimal estimation of $\rho$ value.

### B. Phase 2: Merging of Color Bins

In next phase of extraction, bins whose centroids are near to its neighbor's centroid and bins whose pixel count is less than predefined threshold are merged with its neighboring bins as shown in algorithm 3, this is done to avoid smaller bins that don't contribute as dominant but hold necessary color information. It is important to note that the final bin range obtained is not uniform for all bins. Centroid of bins which is mean of all pixel value present in the bin along the pixel count forms the dominant color pixel values for each channel, the whole process is repeated for remaining two channels.

Fig. 1 and Fig. 2 illustrates the various steps as described above performed over a hypothetical set of pixel value and the end result obtained as dominant color pixel value along with its percentage of occurrence in the set. In Fig. 1.A is shown the pixel values and initial $\rho$ value while in Fig. 1.B is shown the first iteration of binning process. The dotted box indicates non-empty bins, the empty bins are not considered in further iteration of calculation of $\rho$. As the $\rho$ value in further iterations is less than one, the iterative process stops, giving us the results as shown in Fig. 1.C.

Similarly in Fig. 2.A is shows the pixel values and initial $\rho$ value. In Fig. 2.B is shown the first iteration of binning process and dotted box indicate non-empty bins, Fig. 2.C shows the second iteration. The iterative process stops after the second iteration as $\rho$ drops below 1 or no empty bins exists. Fig. 2.D shows the final color bins and associated pixels. It is important to note that the range of pixel value each bins can hold is different and depends on the composition of the image. Finally in Fig. 2.E are given us the dominant color pixel values and their percentage of occurrence.

Algorithm 1: Dominant color pixel value

| |
|---|
| Algorithm : Dominanat Color Pixel value |
| Input : Image of size N*M in RGB color space |
| Output: Dominant color pixel value extractd from each channel namely R,G,B |
| Begin<br>    For Each channel C in Image<br>        //initialize new array of pixel<br>        pixel[]← extract_pixel(Image,C)<br>        sortAscending(pixel[])<br>        Bins[]← saparate(pixel[])<br>        Merge<br>    End For<br>End |

Algorithm 2: Channelized Binning

| |
|---|
| Algorithm: ChannelizedBinning( pixel[])<br>Input: Array of pixels pixel[]<br>Output: Color Bins with pixel in pixel[] classified into them |
| Begin<br>    Compute<br>    Start←pixel[0]<br>    End ← pixel[n]<br>    For i ← 0 to i* End{<br>    Initialize new color Bin $B_i$ with range Start + i*<br>    to Start + (i+1)*<br>    }<br>    Classify pixels in pixel[] into color Bins<br>    j←0<br>    emptyBinFound← false<br>  For each color Bin $B_i${<br>    If($B_i$ *is* empty)<br>    {<br>        emptyBinFound ←true<br>        Copy pixels in $Bin_j$ to $Bin_i$ to Npixel[]<br>        ChannelizedBinning (Npixel[])<br>        Loop till next non-empty Bin $B_i$ is found{<br>            j←i<br>            continue<br>        }<br>    }<br>  }<br>If(emptyBinFound *is* false){<br>    Print the Bin values as output<br>} |

Algorithm 3: Merge Color Bins

| |
|---|
| Algorithm: Merge( Bins[])<br>Input: Array of color bins Bins[]<br>Output: Dominant color pixel value and % of occurrence |
| Begin<br>    For each Bin $B_i$ in Bins[] {<br>        If(number of pixels in Bi < Thresh_count<br>& distance between adjacent Bin $B_j$ < Thresh_distance{<br>            $B_i$←$B_i$+$B_j$<br>        }<br>    }<br>For each Bin $B_i$ in Bins[]{<br>    DominantColorPixelValue$_i$←$Bin_i$ centroid<br>    % of occurrence ←$Bin_i$ pixel count / total number<br>    of pixels<br>    }<br>End |

## III. EXPERIMENTS AND RESULTS

The proposed method was implemented in Open CV workspace. The method was first tested with hand constructed test images whose color composition was known. Later the method was tested against different images of datasets used in [4], [7]. The test images included images of sizes 256*384 and 348*256; the images were compressed using JPEG compression method. Table 1 shows the result of extraction of dominant color pixel values over the constructed test images. The error in the proposed method is estimated using Euclidian distance as in (2).

$$\varepsilon = \frac{\sum_{i=0}^{n} \sqrt{(pixel_{iE} - pixel_{iA})^2 + (percent_{iE} - percent_{iA})^2}}{n} \quad (2)$$

Where $pixel_{iE}$ is the estimated dominant pixel value, $pixel_{iA}$ is the actual dominant pixel value, $percent_{iE}$ is the corresponding estimate percentage of occurrence of $i^{th}$ dominant pixel value and $percent_{iA}$ is actual percentage of occurrence of $i^{th}$ dominant pixel value. Table 2 shows the result of extraction of dominant color pixel value over images chosen from the dataset used in [4], some of the images used for testing are given below.

From the results as shown in table 1, it is possible to conclude that the proposed method of dominant color extraction is successful in extracting the dominant color pixel value, with a Euclidian distance less than 1 from the actual color composition of the image. From the results as shown in table 3, it possible to infer that the proposed methods works satisfactorily for real world images as well and is able to extract 3-5 dominant color pixel value for each color channel.

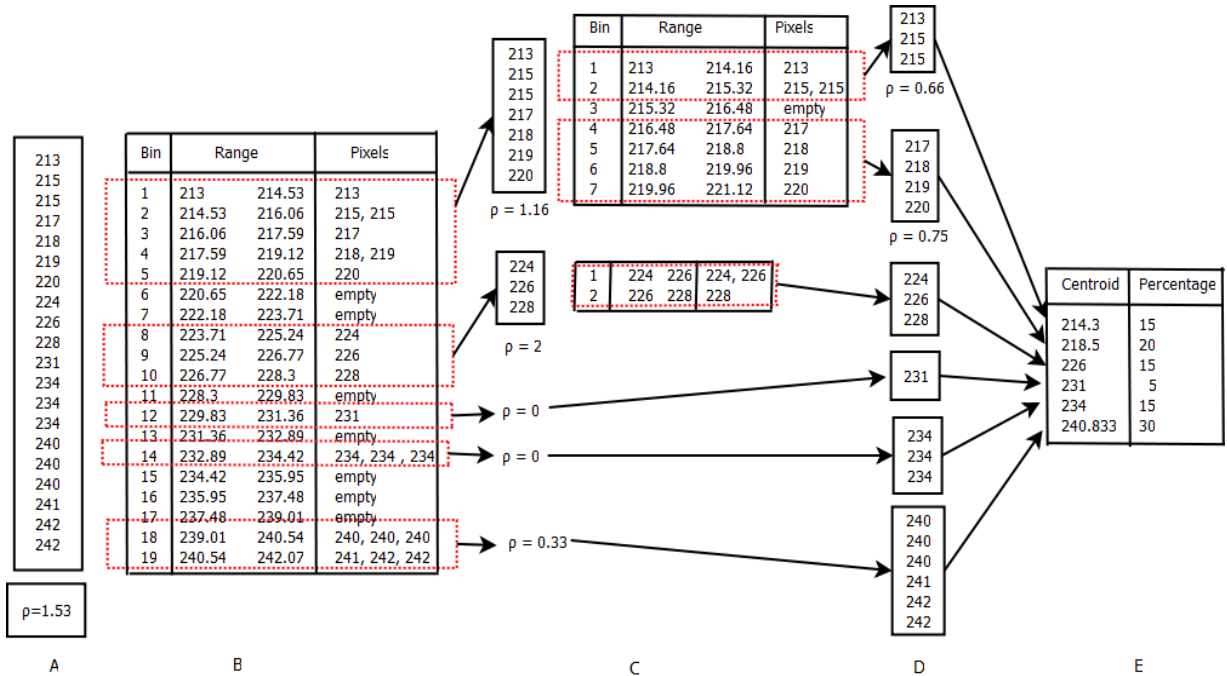

Figure 2: Example 2 calculation on hypothetical pixel values

TABLE I: RESULTS OF DOMINANT COLOR PIXEL VALUE FOR TEST IMAGES

| Image | Estimated Color pixel value By iterative Binning (proposed method) | | | | | | Actual color pixel value | | | | | | Euclidian distance |
|---|---|---|---|---|---|---|---|---|---|---|---|---|---|
| | Red | | Green | | Blue | | Red | | Green | | Blue | | |
| | Pixel value | Percent | Pixel value | Percent | Pixel value | Percent | Pixel value | Percent | Pixel value | Percent | Pixel value | Percent | |
| #1 | 254 | 50.25 | 241 | 50.63 | 0 | 50.38 | 255 | 50.0 | 242 | 50.0 | 0 | 50.0 | 0.3434 |
| | 237 | 49.74 | 27 | 49.36 | 36 | 49.61 | 237 | 50.0 | 28 | 50.0 | 36 | 50.0 | |
| | Red | | Green | | Blue | | Red | | Green | | Blue | | |
| #2 | 0 | 26.57 | 163 | 26.37 | 232 | 27.89 | 0 | 25.2 | 162 | 25.2 | 232 | 25.2 | 0.5367 |
| | 35 | 19.68 | 176 | 19.28 | 77 | 18.84 | 34 | 19.8 | 177 | 19.8 | 76 | 19.8 | |
| | 154 | 5.29 | 217 | 5.46 | 234 | 4.92 | 153 | 7.41 | 217 | 7.41 | 234 | 7.41 | |
| | 164 | 30.54 | 72 | 30.67 | 163 | 30.38 | 163 | 28.2 | 73 | 28.2 | 164 | 28.2 | |
| | 252 | 17.90 | 201 | 18.19 | 13 | 17.95 | 255 | 19.3 | 201 | 19.3 | 14 | 19.3 | |
| | Red | | Green | | Blue | | Red | | Green | | Blue | | |
| #3 | 185 | 5.88 | 122 | 6.10 | 87 | 5.84 | 185 | 8.5 | 122 | 8.5 | 87 | 8.5 | 0.5977 |
| | 154 | 31.65 | 150 | 33.16 | 234 | 33.02 | 153 | 32.2 | 153 | 32.2 | 234 | 32.2 | |
| | 239 | 18.80 | 239 | 18.36 | 175 | 18.24 | 239 | 18.3 | 239 | 18.3 | 176 | 18.3 | |
| | 180 | 43.65 | 228 | 42.36 | 29 | 42.87 | 181 | 40.9 | 230 | 40.9 | 29 | 40.9 | |

TABLE II: CONSTRUCTED IMAGES

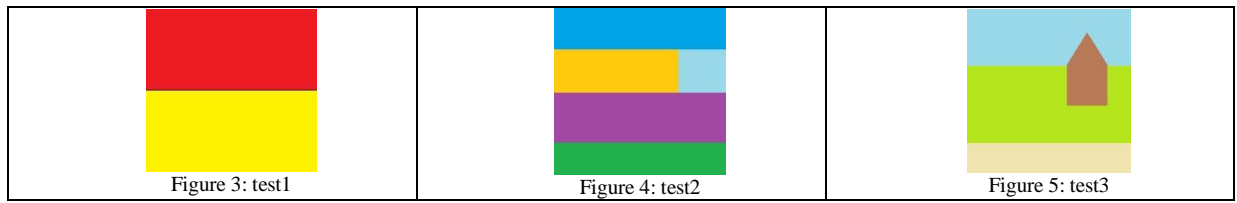

Figure 3: test1    Figure 4: test2    Figure 5: test3

TABLE III: RESULTS OF DOMINANT COLOR PIXEL VALUE FOR IMAGES FROM DATASET

| Image | Red Pixel value | percent | Green Pixel value | percent | Blue Pixel value | percent |
|---|---|---|---|---|---|---|
| #728 | 57 | 59.55 | 74 | 26.38 | 60 | 49.35 |
| | 110 | 28.91 | 138 | 56.10 | 114 | 40.35 |
| | 163 | 10.34 | 188 | 16.43 | 163 | 9.58 |
| | 208 | 1.15 | 228 | 1.073 | 206 | 0.69 |
| #730 | 30 | 60.20 | 32 | 53.75 | 24 | 79.94 |
| | 82 | 26.87 | 92 | 35.88 | 71 | 12.19 |
| | 137 | 6.38 | 141 | 5.85 | 127 | 2.92 |
| | 213 | 6.53 | 206 | 4.49 | 201 | 4.88 |
| #851 | 30 | 40.46 | 31 | 43.28 | 33 | 46.11 |
| | 91 | 33.59 | 88 | 30.63 | 87 | 28.68 |
| | 144 | 9.42 | 140 | 5.32 | 136 | 3.997 |
| | 207 | 16.45 | 216 | 20.74 | 222 | 21.19 |
| #852 | 28 | 37.54 | 29 | 36.62 | 28 | 26.90 |
| | 98 | 14.88 | 100 | 13.47 | 83 | 11.70 |
| | 168 | 39.47 | 174 | 40.37 | 172 | 21.30 |
| | 213 | 8.08 | 215 | 9.52 | 220 | 40.02 |
| #853 | 21 | 51.75 | 27 | 47.30 | 29 | 53.53 |
| | 86 | 37.42 | 102 | 43.62 | 87 | 20.95 |
| | 133 | 5.89 | 152 | 6.02 | 148 | 22.77 |
| | 196 | 4.88 | 199 | 3.04 | 194 | 2.72 |
| #857 | 14 | 39.88 | 19 | 39.39 | 32 | 38.39 |
| | 111 | 38.45 | 79 | 8.013 | 87 | 6.510 |
| | 161 | 11.90 | 151 | 41.85 | 184 | 53.72 |
| | 214 | 9.748 | 205 | 10.71 | 225 | 1.36 |
| #38 | 99 | 78.93 | 107 | 80.06 | 60 | 38.86 |
| | 192 | 21.06 | 197 | 19.93 | 190 | 61.13 |
| #997 | 25 | 5.15 | 31 | 10.15 | 25 | 14.88 |
| | 114 | 21.78 | 106 | 23.78 | 100 | 27.08 |
| | 205 | 71.94 | 175 | 20.91 | 162 | 17.55 |
| | 246 | 1.10 | 223 | 45.14 | 219 | 40.47 |
| #35 | 43 | 51.80 | 46 | 52.18 | 29 | 49.05 |
| | | | | | 240 | 16.02 |
| | 192 | 48.19 | 192 | 47.81 | 162 | 34.92 |
| #26 | 73 | 78.0 | 39 | 44.37 | 25 | 47.79 |
| | 181 | 19.84 | 156 | 50.26 | 202 | 52.202 |
| | 250 | 2.11 | 240 | 5.36 | | |

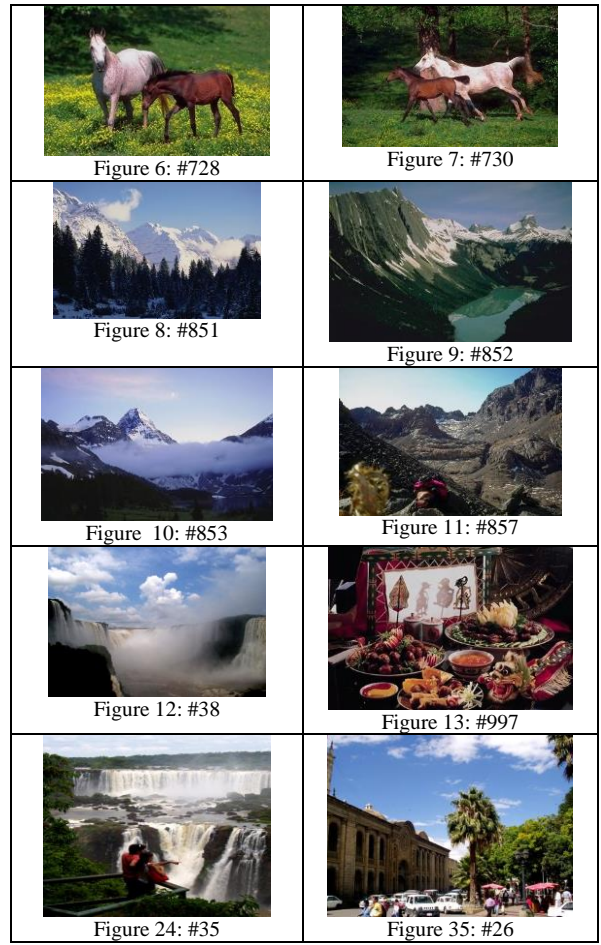

Figure 6: #728 | Figure 7: #730
Figure 8: #851 | Figure 9: #852
Figure 10: #853 | Figure 11: #857
Figure 12: #38 | Figure 13: #997
Figure 24: #35 | Figure 35: #26

## IV. CONCLUSION

In this paper, a novel approach for extraction of dominant color pixel values, which are a variant of the dominant color descriptor, is proposed. The proposed method is shown to perform well with the test and dataset images with the extracts of 3-5 dominant color pixel values for each color channel. The main advantage of the channelized binning method is that it does not make any initial assumption about the number of dominant colors and the centroid of the dominant color. Further it also it operates on 24 bit full RGB color space, by considering one color channel at a time and hence avoiding the color quantization.